# How to Elicit Many Probabilities

**L.C. van der Gaag, S. Renooij,**
Department of Computer Science
Utrecht University
P.O. Box 80.089
3508 TB Utrecht
The Netherlands
{linda, silja}@cs.uu.nl

**C.L.M. Witteman,**
Psychological Laboratory
Utrecht University
P.O. Box 80.140
3508 TC Utrecht
The Netherlands
C.Witteman@fss.uu.nl

**B.M.P. Aleman,** and **B.G. Taal**
The Netherlands Cancer Institute
Antoni van Leeuwenhoekhuis
Plesmanlaan 121
1066 CX Amsterdam
The Netherlands

## Abstract

In building Bayesian belief networks, the elicitation of all probabilities required can be a major obstacle. We learned the extent of this often-cited observation in the construction of the probabilistic part of a complex influence diagram in the field of cancer treatment. Based upon our negative experiences with existing methods, we designed a new method for probability elicitation from domain experts. The method combines various ideas, among which are the ideas of transcribing probabilities and of using a scale with both numerical and verbal anchors for marking assessments. In the construction of the probabilistic part of our influence diagram, the method proved to allow for the elicitation of many probabilities in little time.

## 1 INTRODUCTION

As more and more Bayesian belief networks are being developed for complex problem domains, it is becoming increasingly apparent that the elicitation of all probabilities required is not an easy task. In fact, the elicitation of probabilities is often referred to as a major obstacle in building a Bayesian belief network [Jensen, 1995, Druzdzel & Van der Gaag, 1995]. We experienced the extent to which probability elicitation can be an obstacle to advancement in the construction of the probabilistic part of a complex influence diagram in the field of cancer treatment.

The Antoni van Leeuwenhoekhuis in the Netherlands, hosting the Netherlands Cancer Institute, is specialised in the treatment of cancer patients. In the hospital, every year some hundred patients receive treatment for oesophageal carcinoma. Patients with oesophageal carcinoma currently are assigned to a therapy by means of a standard protocol, involving a small number of variables. Based upon this protocol, 80% of the patients show a favourable response to the therapy instilled. In the context of a project aimed at the development of a more fine-grained protocol with a higher favourable response rate, an influence diagram is being developed for patient-specific therapy selection for oesophageal carcinoma. The influence diagram is destined for use in the Antoni van Leeuwenhoekhuis.

The oesophagus influence diagram is being hand-crafted with the help of two experts in oncology from the Netherlands Cancer Institute. After carefully modeling the characteristics of an oesophageal carcinoma and the possible effects of the various different therapeutic alternatives available in the graphical part of the diagram, we focused on the elicitation of the probabilities required for the diagram's quantitative part. As in many problem domains, various different sources of probabilistic information appeared to be readily available for the elicitation task. Neither data collection nor a thorough literature review, however, yielded any usable results. The single remaining source of probabilistic information, therefore, was the knowledge and personal clinical experience of the two domain experts involved in the project.

For eliciting the conditional probabilities required for the oesophagus influence diagram, we set out using various well-known methods with our domain experts: we used a numerical scale for marking assessments and we used the concept of lotteries [Morgan & Henrion, 1990]. The various problems we encountered with these methods and the amount of time these methods tended to take for the separate assessments, soon revealed that the elicitation of the large number of probabilities required was infeasible with these methods.

Based upon our negative experiences with existing methods, we designed a new method for eliciting probabilities from domain experts. We tailored our method to eliciting a large number of probabilities in little time. As assessments obtained in little time can be quite inaccurate, we envisage the use of our method as the first step in an iterative procedure of stepwise refinement of probability assessments [Coupé et al., 1999]. Our method combines various different ideas. Among these are the ideas of presenting conditional probabilities as fragments of text and of providing a scale for marking assessments with both numerical and verbal anchors. Using our method in the construction of the probabilistic part of the oesophagus influence diagram, we elicited from our domain experts the conditional probabilities required at a rate of 150 – 200 probabilities per hour. In an evaluation interview, the experts indicated that they had felt very comfortable with the method.

The paper is organised as follows. In Section 2, we provide some details of the oesophagus influence diagram and discuss our initial experiences with probability elicitation for the diagram. In Section 3, we describe the method we designed for eliciting a large number of probabilities from domain experts. In Section 4, we evaluate the use of our method in the construction of the probabilistic part of the oesophagus influence diagram; more specifically, we comment on the observations put forward by the domain experts using the method. The paper is rounded off with some conclusions in Section 5.

## 2 THE OESOPHAGUS INFLUENCE DIAGRAM

As a consequence of a lesion of the oesophageal wall, for example as a result of frequent reflux, a carcinoma may develop in a patient's oesophagus. An oesophageal carcinoma has various characteristics that influence its prospective growth. These characteristics include the location of the carcinoma in the oesophagus, the histological type of the carcinoma, its length, and its macroscopic shape. An oesophageal carcinoma typically invades the oesophageal wall and upon further growth may invade neighbouring structures such as the trachea and bronchi. In due time, the carcinoma may give rise to lymphatic metastases in distant lymph nodes and to haematogenous metastases in, for example, the lungs and the liver of the patient. The depth of invasion and the extent of the metastases of the carcinoma largely influence a patient's life expectancy.

While establishing the presence of an oesophageal carcinoma in a patient is relatively easy, the selection of an appropriate therapy is a far harder task. In the Antoni van Leeuwenhoekhuis, various different therapeutic alternatives are available, ranging from surgical removal of the oesophagus, to radiotherapy, and positioning a prosthesis in the oesophagus. The effects aimed at by instilling a therapy include removal or reduction of the patient's primary tumour and an improved passage of food through the oesophagus. The various therapeutic alternatives available differ in the extent to which these effects can be attained. Instillation of a therapy further is expected to be accompanied not just by beneficial effects but also by various complications; these complications can be very serious and may even lead to death. The effects and complications to be expected from the various therapeutic alternatives available for a patient depend on the characteristics of his or her carcinoma, on the depth of invasion of the carcinoma into the oesophageal wall and neighbouring structures, and on the extent of the carcinoma's metastases. It will be evident that the possible effects and complications require careful balancing before a therapy is decided upon.

The overall structure of the oesophagus influence diagram is shown in Figure 1. The graphical part of the diagram was handcrafted with the help of two domain experts from the Netherlands Cancer Institute; the construction of this graphical part took approximately two years, with one two-hour interview every two or three weeks. The influence diagram currently includes one decision node and over 70 chance nodes. Of these, 40 chance nodes pertain to the characteristics of an oesophageal carcinoma, to the depth of its invasion, and to the extent of its metastases; the remaining chance nodes model the possible effects and complications of the various therapies available. For the chance nodes, a total of almost 3000 probabilities is required.

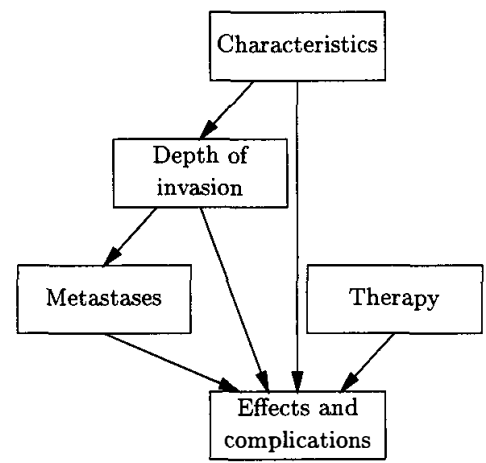

Figure 1: The Overall Structure of the Oesophagus Influence Diagram.

So far, we focused our elicitation efforts on the part of the diagram that pertains to the characteristics, depth of invasion, and metastases of an oesophageal carcinoma. This part constitutes a coherent and self-supporting Bayesian belief network. The 40 nodes in-volved require some thousand probability assessments. The node requiring the largest number of assessments, 144, models the staging of the carcinoma; this node is a deterministic node, classifying an oesophageal carcinoma according to the depth of its invasion and the extent of its metastases. The non-deterministic node requiring the largest number of probability assessments is the node describing the result of an echo-endoscopic examination of a patient's oesophagus with respect to the depth of invasion of the carcinoma in the oesophageal wall; it requires 80 assessments.

The elicitation of the probabilities required for the 40 nodes indicated above proved to be a major obstacle in the construction of our influence diagram. As in many problem domains, various sources of probabilistic information were available. We collected data from historical patient records and we performed a literature review. Unfortunately, the Netherlands being a low-incidence country for oesophageal carcinoma, we were not able to compose an up-to-date, large and rich enough data collection to allow for reliable assessment of the probabilities required; after due consideration, we decided instead to save the collected data for evaluation purposes. Our literature review, although very thorough, also did not result in ready-made assessments; in fact, hardly any results reported in the literature turned out to be usable for our influence diagram. The single remaining source of probabilistic information, therefore, was the knowledge and personal clinical experience of the two domain experts involved in the project. The problems of bias and poor calibration that, unfortunately, are typically encountered when eliciting judgemental probabilities from experts are widely known [Kahneman et al., 1982].

In our first efforts to elicit conditional probabilities for the oesophagus influence diagram from our domain experts, we focused on the use of a probability scale for marking assessments, on the frequency method, and on the concept of lotteries, as well-known methods for probability elicitation [Morgan & Henrion, 1990, Gigerenzer & Hoffrage, 1995]. These methods were designed to avert to at least some extent the typical problems found with human probability assessment. Before commenting on our experiences with these methods, we would like to emphasise that, prior to the construction of the oesophagus influence diagram, the domain experts involved had little or no acquaintance of expressing their knowledge and clinical experience into numbers.

The probability scale we used in the elicitation was a horizontal line with the three anchors 0, 0.5, and 1. The domain experts were asked to mark their assessments for *all* conditional probabilities pertaining to a single variable given some conditioning context on the same line. The probabilities to be assessed were presented to them in mathematical notation. Unfortunately, the experts involved in the project had some difficulties working with the mathematical notation. In addition, they felt quite uncomfortable with the probability scale, as it gave them very little to go by. The request to mark various assessments on a single line further appeared to introduce a bias towards aesthetically distributed marks.

With the frequency method, the domain experts were asked to envisage a population of one hundred patients suffering from an oesophageal carcinoma with certain characteristics. They were asked to assess the number of patients from among this population who would show a characteristic under study. Since oesophageal carcinoma has a low incidence in the Netherlands, visualising one hundred patients with certain specific characteristics turned out to be a demanding, if not impossible, task.

The use of lotteries for probability elicitation, unfortunately, also entailed various difficulties. The domain experts indicated that they often felt confronted with lotteries that were very hard to conceive because of the rare, or unethical, situations they represented. Moreover, the use of lotteries tended to take so much time that it soon became apparent that the elicitation of several thousands of conditional probabilities in this way was quite infeasible.

## 3 THE ELICITATION METHOD

For the probabilistic part of the oesophagus influence diagram, several thousands of conditional probabilities had to be assessed. As we argued in the previous section, these probabilities had to be elicited from the domain experts involved in the construction of the diagram. Experience with well-known methods for probability elicitation had shown that assessing all probabilities required was no easy task. Based upon the negative experiences with these methods, we designed a new method for eliciting probabilities from domain experts that is tailored to the elicitation of a large number of conditional probabilities in little time. In this section, we present our method; its use will be commented upon in Section 4.

Our method for probability elicitation from domain experts combines various different ideas. Although several of these ideas were presented before by others, we combined and enhanced them to yield a novel and, as we will argue in the next section, effective elicitation method. At the heart of our method lies the idea of presenting domain experts with a separate figure for every conditional probability that needs to be assessed. Figure 2 shows, as an example, the figure pertaining

Consider a patient with a *polypoid* oesophageal carcinoma; the carcinoma has a length of *less than 5 cm*. How likely is it that this carcinoma invades into the *muscularis propria (T2)* of the patient's oesophageal wall, but not beyond ?

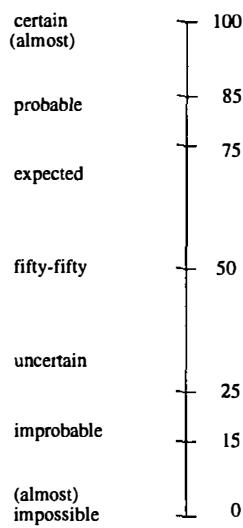

Figure 2: The Fragment of Text and Probability Scale for the Assessment of the Conditional Probability $\Pr(Invasion = T2 \mid Shape = polypoid, Length < 5\ cm)$.

to the conditional probability

$$\Pr(\ Invasion = T2 \quad \mid \quad Shape = polypoid, Length < 5\ cm)$$

for the oesophagus influence diagram. On the left of the figure is a fragment of text that transcribes the conditional probability to be assessed. Using a fragment of text to denote a probability circumvents the need to use mathematical notation. The fragment is stated in terms of likelihood rather than in terms of frequency to forestall difficulties with the assessment of a conditional probability for which the conditioning context is quite rare. To facilitate the assessment of a required probability, a vertical scale is depicted to the right of the text fragment. Indicated on this scale are various different numerical and verbal anchors. We will presently comment on the specific anchors used.

The figures pertaining to the various conditional probabilities to be assessed are grouped in such a way that the probabilities from the same conditional distribution can be taken into consideration simultaneously; the figures are presented in groups of two or three on consecutive single-sided sheets of paper. Explicitly grouping related probabilities has the advantage of reducing the number of times a mental switch of conditioning context is required of the domain experts during the elicitation.

The probability scale to be used with our method specifies both numerical and verbal anchors. Research on human probability judgement has indicated that most people tend to feel more at ease with verbal probability expressions than with numbers. Verbal expressions of probability are considered to be more natural than numerical probabilities, easier to understand and communicate, and better suited to convey the vagueness of one's opinions [Wallsten et al., 1993]; this observation also holds for physicians and other health workers [Merz et al., 1991]. Words, however, should not self-evidently be preferred to numbers, as neither should numbers to words. In fact, the two modes of communicating probabilistic information can both be used [Brun & Teigen, 1988]. Motivated by these observations, we decided to search for a probability scale for marking assessments that is based on *both* numbers *and* verbal probability expressions.

To develop a scale of verbal probability expressions to be used with numbers, we undertook four separate studies. In the first study, we asked subjects to provide a list of the verbal probability expressions they commonly use. This study yielded seven most frequently used expressions, being (translated from the corresponding Dutch expressions) "certain", "probable", "expected", "fifty-fifty", "uncertain", "improbable", and "impossible". In the second study, (other) subjects were asked to rank order these expressions. The results from this study indicated that the seven verbal probability expressions had a considerably stable rank ordering between subjects. To establish the relative distances among the seven expressions, in the third study, subjects were asked to compare each pair of expressions and assess the degree to which the two expressions conveyed the same probability. The distances yielded by this study were used to project the verbal probability expressions onto a numerical scale. The expression "certain" was calculated to be equivalent to 100%, "probable" to be equivalent to approximately 85%, and "expected" approximately to 75%; "fifty-fifty" was calculated to be equal to 50%, "uncer-tain" approximately to 25%, "improbable" to approximately 15%, and "impossible", to conclude, was calculated to denote 0%. Using this projection of verbal probability expressions onto numbers, the fourth study focused on the question whether decisions were influenced by the mode in which probability information was presented. The results indicated that the decisions made were independent of whether the probability information was expressed numerically or verbally. We would like to note that the four studies included subjects from the field of medicine. For further details of the studies, we refer the reader to an extended paper [Renooij & Witteman, 1999].

The fact that the subjects in our studies interpreted the verbal probability expressions as intended, motivated us to further elaborate on a scale with both numerical and verbal anchors for use as an aid for probability elicitation. Since the verbal probability expressions were explicitly intended as independent anchors on the scale rather than as translations for the numerical probabilities, we decided to position the verbal probability expressions close by rather than simply beside the numerical anchors. We further decided to add the moderator "(almost)" to the most extreme verbal expressions to indicate the positions of very small and very large probabilities. The resulting probability scale is the scale shown in Figure 2.

## 4 EVALUATION OF THE ELICITATION METHOD

We used our newly designed method for probability elicitation from domain experts in the construction of the probabilistic part of the oesophagus influence diagram. In this section, we evaluate the use of our method. More specifically, we comment upon the observations put forward by the domain experts involved. In addition, we briefly review the preliminary results from an initial evaluation of the influence diagram in the making.

### 4.1 USING THE METHOD

The elicitation of the conditional probabilities required for the part of the oesophagus influence diagram outlined in Section 2, took approximately two months, with one two-hour interview with the domain experts every two weeks. Each interview focused on a small coherent part of the diagram. Prior to every interview, the elicitors spent some ten hours preparing the fragments of text to be presented with our probability scale to the experts.

In the first interview, the domain experts were informed of the basic ideas underlying the new elicitation method. The intended use of the probability scale was detailed to them. In addition, the experts were told that their initial probability assessments would be subjected to a sensitivity analysis to reveal the sensitivities of the diagram's results to the various assessments, and that we would try and refine the most influential ones later on in the project; for details of our procedure for stepwise refinement of assessments, we refer the reader once again to [Coupé et al., 1999]. The basic idea of sensitivity analysis was explained to the domain experts in depth to reassure them that rough assessments for the requested conditional probabilities would suffice at this stage in the construction of the influence diagram.

Following the last interview, the domain experts were asked to evaluate the new method we had used with them for probability elicitation. For this purpose, a written evaluation form was designed so as to not influence their observations. In the evaluation, the domain experts indicated that they had felt very comfortable with the elicitation method. They found the method most effective and much easier to use than any method for probability elicitation they had used before.

We recall from Section 3 that one of the ideas underlying our method for probability elicitation is the use of a fragment of text to indicate a conditional probability that needs to be assessed. The use of these fragments of text seemed to work very well. The two domain experts mentioned that they had had no difficulties understanding the described patient characteristics. During the interviews, the elicitors had often noted that the described characteristics served to call to mind various different concrete patients. Although the experts could not envisage a large group of patients with certain specific characteristics, their extensive clinical experience with cancer patients in general and their knowledge of reactive growth of cancer cells, along with information recalled from literature, enabled them to provide the requested assessments.

With respect to the probability scale used for marking assessments, the domain experts indicated that they had found the presence of both numerical and verbal anchors helpful. They mentioned that, upon thinking about a conditional probability to be assessed, they used words as well as numbers. Depending on how familiar they felt with the characteristics described in a fragment of text, they preferred using verbal or numerical expressions. The more uncertain they were about the probability to be assessed, for example, the more they were inclined to think in terms of words. The verbal anchors on the scale then helped them to determine the position that they felt expressed the probability they had in mind.

The two domain experts further mentioned that they had felt comfortable with the specific verbal anchors used with the probability scale. They indicated, however, that the expression "impossible" is hardly ever used in oncology. Especially in their communication with patients, oncologists appear to prefer the more cautious expression "improbable" to refer to almost impossible events. As a consequence, our domain experts tended to interpret the expression "improbable" as a 5% or even smaller probability rather than as a probability of around 15%. Since the probability scale provided both words and numbers, they had no difficulties indicating what they meant to express. The experts also mentioned that an extra anchor around 40% would have been useful. Note that these observations pertain to the lower half of the scale only. We would like to add to these observations that our probability scale hardly accommodates for indicating extreme probability assessments, that is, assessments very close to 0 or 1. During the various interviews, however, the domain experts never seemed to want to express such extreme assessments. When asked about extreme probabilities, they confirmed our observation.

Another idea underlying our method is the idea of grouping the figures used in such a way that the probabilities from the same conditional distribution can be taken into consideration simultaneously. During the elicitation interviews, the domain experts were advised first to focus on the probabilities from a conditional distribution that were the easiest to assess, and then to distribute the remaining probability mass over the more difficult probabilities. This turned out to be a most effective heuristic for eliciting assessments for variables with more than two or three values.

To conclude, we would like to point out that, during the earlier, rather unsuccessful, elicitation efforts, our domain experts had acquired some acquaintance of expressing their knowledge and personal clinical experience into numbers. As a result, they now appeared to be less daunted by the assessment task.

### 4.2 THE USE OF TRENDS

During the elicitation interviews with our domain experts, the concept of *trend* emerged. We use the term to denote a fixed relation between two conditional probability distributions for the same variable given different conditioning contexts.

To illustrate the concept of trend, we address the variable *Invasion* modelling the depth of invasion of an oesophageal carcinoma into the wall of a patient's oesophagus. This variable can take one of the values *T1*, *T2*, *T3*, and *T4*; the higher the number indicated in the value, the deeper the carcinoma has invaded into the oesophageal wall and the worse off the patient is. For the variable *Invasion*, various conditional probabilities had to be assessed, pertaining to differing shapes and varying lengths of a carcinoma. Upon assessing the conditional probabilities required for the variable *Invasion*, the domain experts started with the probabilities for the depth of invasion of a *polypoid* oesophageal carcinoma with a length of less than 5 centimeters. They subsequently indicated that patients with *ulcerating* tumours of this length were 10% worse off with regard to the depth of invasion of the carcinoma than patients with equivalent polypoid tumours. They thus explicitly related two conditional probability distributions for the variable *Invasion* to one another. As trends appeared to be a quite natural way of expressing probabilistic information, we encouraged our domain experts to provide trends wherever appropriate.

We designed a generic method for handling the trends provided by our domain experts in an intuitively appealing and mathematically correct way. The method is best explained in terms of the example trend stated above. Suppose that, given a *polypoid* oesophageal carcinoma of less than 5 centimeters, the probabilities for the four different values of the variable *Invasion* have been assessed at $x_1$, $x_2$, $x_3$, and $x_4$ — $x_i$ being the probability assessment for the value $Ti$. After consultation with our domain experts, we interpreted the specified trend as follows: *10% of the patients with a polypoid tumour of less than 5 centimeters with $Ti$ for its depth of invasion would have had $Ti + 1$ for the depth of invasion if the tumour would have been an ulcerating tumour, $i = 1, 2, 3$.* The basic idea of the interpretation of the trend is depicted in Figure 3. For the probability assessments $y_1$, $y_2$, $y_3$, and $y_4$ for the different values of the variable *Invasion* given an *ulcerating* oesophageal carcinoma of less than 5 centimeters, we thus find

$$y_1 \leftarrow x_1 - 0.10 \cdot x_1$$
$$y_2 \leftarrow x_2 - 0.10 \cdot x_2 + 0.10 \cdot x_1$$
$$y_3 \leftarrow x_3 - 0.10 \cdot x_3 + 0.10 \cdot x_2$$
$$y_4 \leftarrow x_4 + 0.10 \cdot x_3$$

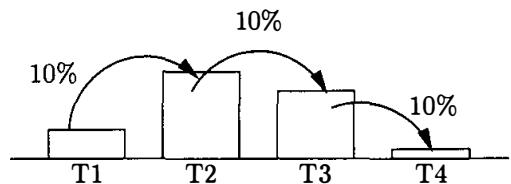

Figure 3: Handling Trends.

It is readily verified that the resulting assessments $y_1$, $y_2$, $y_3$, and $y_4$ each lie between 0 and 1, and together add up to 1. In addition, it will be evident that this method for handling trends can easily be generalised to variables with another number of values and to trends specifying other percentages and other directions of change.

### 4.3 AN INITIAL EVALUATION OF THE DIAGRAM

With our new method, we elicited from the domain experts involved all conditional probabilities required for the part of the oesophagus influence diagram that pertains to the characteristics, depth of invasion, and metastases of an oesophageal carcinoma. As mentioned before in Section 2, this part of the diagram constitutes a coherent and self-supporting Bayesian belief network; it provides for predicting the stage of a patient's oesophageal carcinoma from the results of various different diagnostic tests. To get some preliminary insight in the quality of the influence diagram in the making, we performed an initial evaluation of the 40-node belief network with patient data from 184 patients, available from the Netherlands Cancer Institute. Before detailing this evaluation and its results, we would like to note that the data collection used is known to be biased, to contain inconsistencies, and to be incomplete in a non-random way.

For each patient from our data collection, we instantiated, in the belief network, all nodes pertaining to diagnostic tests for which a test result was available for the patient under consideration. These diagnostic tests range from a biopsy of the primary tumour to an echo-endoscopic examination of the oesophagus, and an X-ray of the patient's chest. From the thus partially instantiated belief network, we computed the most likely stage of the patient's oesophageal carcinoma and compared it with the stage recorded in the data. The stage of an oesophageal carcinoma can be either I, IIA, IIB, III, IVA, or IVB.

For 29 patients from our data collection, unfortunately, the stage of the oesophageal carcinoma was not recorded, which left us with 155 patients for our evaluation. In 94 of these 155 patients, the stage of the carcinoma recorded in the data matched the stage with highest probability computed from the belief network. Under the assumption that the stages recorded in the data are correct, therefore, in 61% of the patients the network predicted the correct stage. We would like to note that this percentage is not uncommon in evaluations of knowledge-based systems [Berner et al., 1994].

Careful examination of the data of the patients for which the belief network returned an incorrect stage learned that the network's prediction deviated from the data most notably for patients with an oesophageal carcinoma of stage IVB. For some 70% of the patients with a IVB-staged carcinoma, another stage was predicted by the network; the stage IVB was quite often yielded as the second most likely stage, however. After removing all patients with a IVB-staged carcinoma from our data collection, the network predicted the correct stage for 68% of the remaining patients.

To conclude our initial evaluation, we re-addressed the data of patients for whose oesophageal carcinoma the belief network predicted a stage that differed from the stage recorded in the data; in doing so, we once again included the patients with a IVB-staged carcinoma. Since most probability assessments for the network had been rounded off at 5%, we investigated the effect, on the percentage of correct predictions, of considering certain stages as (almost) correct. To this end, we considered an oesophageal carcinoma as (almost) correctly staged by the network, if for the stage recorded in the data a probability was computed from the network that differed by at most 5% from the probability of the most likely stage. The percentage of correct predictions then approached 70%. Given that the probabilities used are rough, initial assessments and that the patient data definitely require clearing out, the results from the initial evaluation are quite encouraging.

## 5  CONCLUSIONS

We experienced the extent to which probability elicitation can be an obstacle to advancement in the construction of the probabilistic part of the oesophagus influence diagram. Motivated by our negative experiences with existing methods, we designed a new method for eliciting probabilities from domain experts. Our method combines various different ideas, among which are the ideas of transcribing probabilities and of using a scale with both numerical and verbal anchors. We used our new elicitation method for eliciting the probabilities required for the oesophagus influence diagram and evaluated its use with the domain experts involved. The experts indicated that they found the method much easier to use than any method for probability elicitation they had used before.

For the construction of the oesophagus influence diagram, our newly designed elicitation method entailed a major breakthrough. Prior to the use of our method, we had spent over a year experimenting, on and off, with other methods for probability elicitation without success. Using our elicitation method, the probabilities for a major part of the oesophagus influence diagram were elicited in just two months' time. Our method tends to take considerable time on the part of the elicitors in preparing the various interviews with the experts. However, the ease with which probabili-ties are elicited with the method makes this time certainly well spent.

## References

[Berner et al., 1994] E.S. Berner, G.D. Webster, A.A. Shugerman, J.R. Jackson, J. Algina, A.L. Baker, E.V. Ball, C.G. Cobbs, V.W. Dennis, E.P. Frenkel, L.D. Hudson, E.L. Mancall, C.E. Rackley, and O.D. Taunton (1994). Performance of four computer-based diagnostic systems. *The New England Journal of Medicine*, vol. 330, pp. 1792 – 1796.

[Brun & Teigen, 1988] W. Brun and K.H. Teigen (1988). Verbal probabilities: ambiguous, context-dependent, or both ? *Organizational Behavior and Human Decision Processes*, vol. 41, pp. 390 – 404.

[Coupé et al., 1999] V.M.H. Coupé, L.C. van der Gaag, and J.D.F. Habbema (1999). Sensitivity analysis: an aid for belief-network quantification, submitted for publication.

[Druzdzel & Van der Gaag, 1995] M.J. Druzdzel and L.C. van der Gaag (1995). Elicitation of probabilities for belief networks: combining qualitative and quantitative information. *Proceedings of the Eleventh Conference on Uncertainty in Artificial Intelligence*, pp. 141 – 148.

[Gigerenzer & Hoffrage, 1995] G. Gigerenzer and U. Hoffrage (1995). How to improve Bayesian reasoning without instruction: frequency formats. *Psychological Review*, vol. 102, pp. 684 – 704.

[Jensen, 1995] A.L. Jensen (1995). Quantification experience of a DSS for mildew management in winter wheat. In: M.J. Druzdzel, L.C. van der Gaag, M. Henrion, and F.V. Jensen. *Working Notes of the Workshop on Building Probabilistic Networks: Where Do the Numbers Come From ?*, pp. 23 –31.

[Kahneman et al., 1982] D. Kahneman, P. Slovic, A. Tversky (1982). *Judgment under Uncertainty: Heuristics and Biases*. Cambridge University Press.

[Merz et al., 1991] J.F. Merz, M.J. Druzdzel, and D.J. Mazur (1991). Verbal expressions of probability in informed consent litigation. *Medical Decision Making*, vol. 11, pp. 273 – 281.

[Morgan & Henrion, 1990] M.G. Morgan, M. Henrion (1990). *Uncertainty, a Guide to Dealing with Uncertainty in Quantitative Risk and Policy Analysis*. Cambridge University Press, Cambridge.

[Renooij & Witteman, 1999] S. Renooij, C.L.M. Witteman (1999). Talking probabilities: communicating probabilistic information with words and numbers, submitted for publication.

[Wallsten et al., 1993] T.S. Wallsten, D.V. Budescu, R. Zwick (1993). Comparing the calibration and coherence of numerical and verbal probability judgments. *Management Science*, vol. 39 (2), pp. 176 – 190.